\documentclass[10pt,twocolumn,letterpaper]{article}

\usepackage[pagenumbers]{cvpr} 

\usepackage{graphicx}
\usepackage{amsmath}
\usepackage{amssymb}
\usepackage{booktabs}

\usepackage[accsupp]{axessibility} 
\usepackage[table]{xcolor}
\usepackage{etoolbox}
\usepackage{tabularx}

\usepackage{times}
\usepackage{epsfig}
\usepackage{graphicx}
\usepackage{verbatim}
\usepackage{hhline}
\usepackage{multirow}
\usepackage{makecell}
\usepackage{xcolor}
\usepackage{pifont} 
\usepackage{xspace}

\definecolor{gray}{rgb}{0.5, 0.5, 0.5}
\definecolor{darkgray}{rgb}{0.2, 0.2, 0.2}
\definecolor{scarlet}{rgb}{1, 0.35, 0.15}
\definecolor{mulberry}{rgb}{0.773, 0.294, 0.549}
\definecolor{blue}{rgb}{0, 0, 1}
\definecolor{skyblue}{rgb}{0.3, 0.7, 0.9}
\definecolor{darkgreen}{rgb}{0.2, 0.7, 0.1}
\definecolor{darkyellow}{rgb}{1, 0.65, 0}
\definecolor{lightgray}{rgb}{0.9, 0.9, 9}

\usepackage[pagebackref,breaklinks,colorlinks]{hyperref}

\usepackage[capitalize]{cleveref}

\newcommand{\Tref}[1]{Table~\textcolor{blue}{\ref{#1}}}

\newcommand{\Eref}[1]{Eq.~\textcolor{black}{(\ref{#1}})}
\newcommand{\Erefs}[2]{Eqs.~\textcolor{black}{\eqref{#1}}--\eqref{#2}}
\newcommand{\Fref}[1]{Fig.~\textcolor{blue}{\ref{#1}}}
\newcommand{\Sref}[1]{Sec.~\textcolor{blue}{\ref{#1}}}

\definecolor{darkgreen}{RGB}{0,127,0}
\definecolor{darkred}{RGB}{200,0,0}
\def\greencheckmark{\textcolor{darkgreen}{\checkmark}}
\def\redxmark{\textcolor{darkred}{\ding{55}}}  

\def\cvprPaperID{476} 
\def\confName{CVPR}
\def\confYear{2023} 

\newcommand{\link}[1]{{\color{blue}\href{#1}{#1}}}

\begin{document}

\title{TTA-COPE: Test-Time Adaptation for Category-Level Object Pose Estimation}

\author{Taeyeop Lee$^{1}$ \quad Jonathan Tremblay$^{2}$ \quad Valts Blukis$^{2}$ \quad Bowen Wen$^{2}$ \quad Byeong-Uk Lee$^{1}$ \\
Inkyu Shin$^{1}$ \quad Stan Birchfield$^{2}$ \quad In So Kweon$^{1}$ \quad Kuk-Jin Yoon$^{1}$\\ 
$^{1}$KAIST~~~~~~$^{2}$NVIDIA~~
}

\maketitle

\begin{abstract}
Test-time adaptation methods have been gaining attention recently as a practical solution for addressing source-to-target domain gaps by gradually updating the model without requiring labels on the target data. In this paper, we propose a method of test-time adaptation for category-level object pose estimation called TTA-COPE. We design a pose ensemble approach with a self-training loss using pose-aware confidence. Unlike previous unsupervised domain adaptation methods for category-level object pose estimation, our approach processes the test data in a sequential, online manner, and it does not require access to the source domain at runtime. Extensive experimental results demonstrate that the proposed pose ensemble and the self-training loss improve category-level object pose performance during test time under both semi-supervised and unsupervised settings. Project page: \link{https://taeyeop.com/ttacope}
\end{abstract}

\section{Introduction}
Object pose estimation is a crucial problem in computer vision and robotics.
Advanced methods that focus on diverse variations of object 6D pose estimation have been introduced, such as known 3D objects (instance-level)~\cite{tremblay2018corl:dope, peng2019pvnet}, category-level~\cite{wang2019normalized, Tian2020prior,lin2022icra:centerpose}, few-shot~\cite{he2022fs6d}, and zero-shot pose estimation~\cite{labbemegapose,wen2021bundletrack}.
These techniques are useful for downstream applications requiring an online operation, such as robotic manipulation~\cite{mousavian20196, wen2022you, gao2021kpam} and augmented reality~\cite{runz2018maskfusion, marchand2015pose, marder2016project}.
Our paper focuses on the category-level object pose estimation problem since it is more broadly applicable than the instance-level problem. 

\begin{figure}[t]
\begin{center}
\includegraphics[width=1\linewidth]{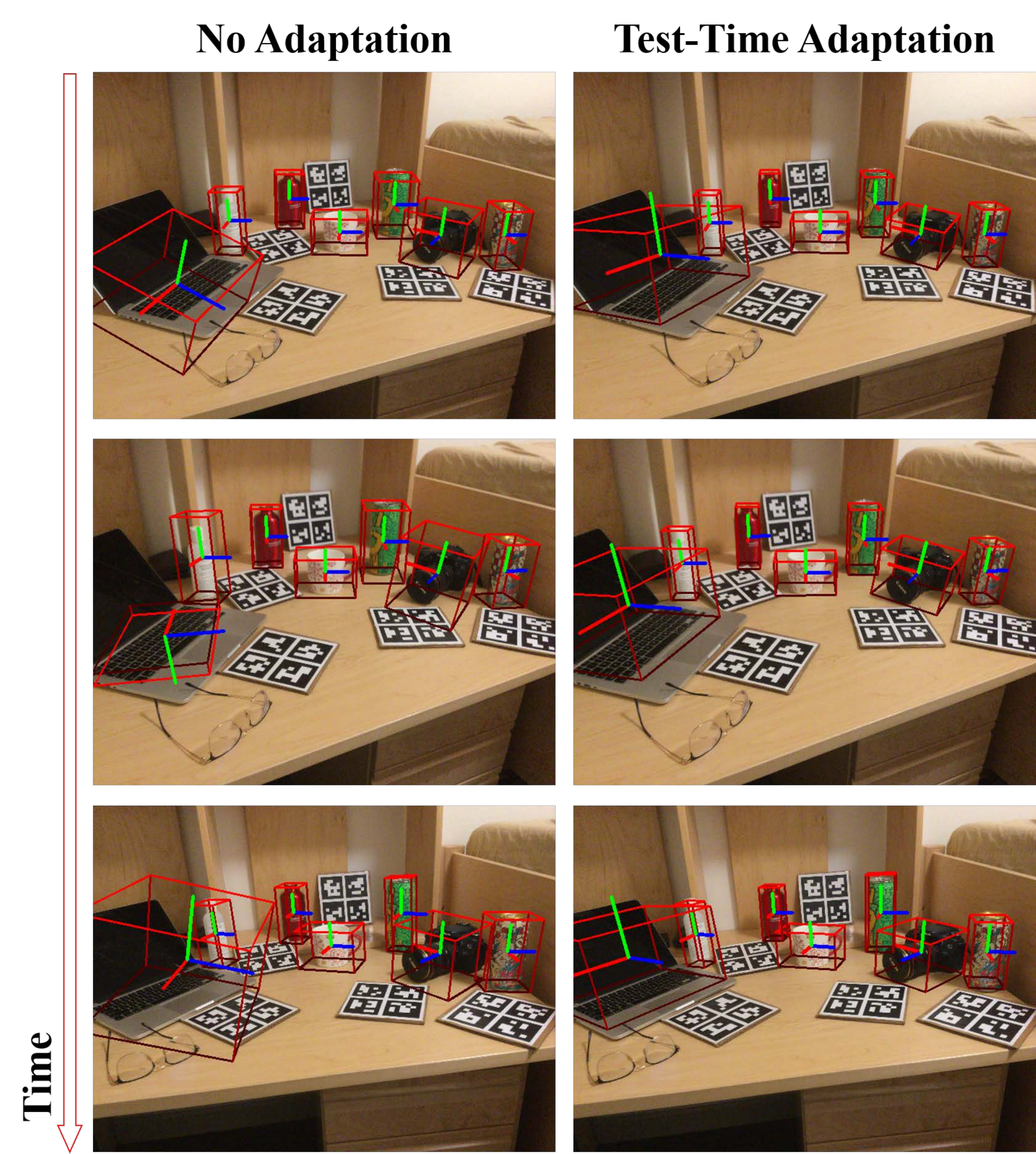}
\vspace{-0.25in}
\caption{We propose a Test-Time Adaptation for Category-level Object Pose Estimation framework (TTA-COPE) that automatically improves the network in an online manner without labeled target data.
As new image frames are processed, our method fine-tunes the network using the unlabeled data and simultaneously applies the network to perform pose estimation via inference.
This approach successfully handles domain shifts compared with no adaptation, as seen here.
}
\label{fig:teaser}
\end{center}
\vspace{-0.25in}
\end{figure} 

Many works on category-level object pose estimation~\cite{wang2019normalized, Tian2020prior,  chen2020cass, lin2021dualposenet, wang2021category, chen2021sgpa,lin2022icra:centerpose} have been proposed recently. 
These approaches estimate multiple classes of object pose more efficiently in a single network compared to the instance-level object pose estimation methods~\cite{tremblay2018corl:dope, xiang2017posecnn, park2019pix2pose,  wang2019densefusion, wen2020se, wen2020robust}, which depend on known 3D shape knowledge and the size of the objects.
Notably, Wang~\etal~\cite{wang2019normalized} introduced a novel representation called Normalized Object Coordinate Space (NOCS) to align various object instances within each category in a canonical 3D space. 
The strengths of the NOCS representation have led to its adoption by follow-up work~\cite{Tian2020prior, lin2021dualposenet, chen2021sgpa}. 

In order to obtain accurate category-level object pose methods in unseen real-world scenarios, it is desirable to fine-tune the models in the new environment with labeled target data.
The model that is not fine-tuned on the target domain distribution will almost certainly exhibit lower performance than the fine-tuned model~\cite{tremblay2018training}.
However, annotating 6D poses of objects in the target environment is an  expensive process~\cite{wang2019normalized, ahmadyan2021objectron, tyree20226, wang2022phocal} that we seek to avoid.
Compared to annotating in 2D space, labeling in 3D space requires specific knowledge about geometry~\cite{hartley2003multiple} from the annotator and is much more laborious, time-consuming, and error-prone due to the complex nature of $SE(3)$ space.
Therefore, it is usually challenging to annotate real-world data with 3D annotations for fine-tuning.

In order to solve the aforementioned problem of annotating object pose data in the real world, several recent methods~\cite{lee2022uda, lin2022category, fu2022category} propose unsupervised domain adaptation (UDA) that aims to train the network without utilizing the ground truth of target pose labels.
Although they show promising results using UDA techniques, these approaches still do not meet some of the requirements for online applications.
For example, when a robot encounters a new environment, it is desirable to adapt the scene online manner while estimating object poses rather than waiting for enough data to be collected in the novel scene to train the model offline.

This problem definition of online fine-tuning is more practical for real applications, where we desire to update the model instantly when new data becomes available for fast domain adaptation.
This setting is known as test-time adaptation (TTA)~\cite{wang2021tent}.
For TTA, the requirements are as follows:
1) labeled source data should not be accessed at test time, 2) adaptation should be online (rather than offline batch processing), and 3) the method should be fully unsupervised, without using 2D or 3D target labels during online fine-tuning.
Since we do not have access to labeled source data (source-free) at test time this problem is more challenging than existing unsupervised category-level object pose methods~\cite{lee2022uda, lin2022category, fu2022category, peng2022self}. 
\Tref{tab:related_work} summarizes the difference between our problem definition and existing methods, showing that test-time adaptation for category-level object pose estimation remains an open problem.

\begin{table}[t]
\centering
\caption{\textbf{Comparison with prior unsupervised works for category-level object pose estimation.} 
Our unsupervised method trains models without 2D or 3D labels of target data, similar to Self-DPDN~\cite{lin2022category}. Unlike previous methods, our proposed approach updates the model online without offline batch processing. 
Moreover, we do not use the source data during test time (source-free) because it is impractical to train on a large amount of source data every iteration. There also may be privacy or legal constraints to access source data~\cite{liu2021source}.
} \vspace{-10pt}
    \resizebox{1.0\linewidth}{!}{
\begin{tabular}{ccc c cc}
\\ \Xhline{4\arrayrulewidth}
           \multirow{3}{*}{Method} & \multicolumn{2}{c}{\textbf{Unsupervised}}                                                                                & & \multicolumn{2}{c}{\textbf{Test-time Adaptation}} \\ \cline{2-3} \cline{5-6}
           & Target 3D & Target 2D & & Source-Free & \begin{tabular}[c]{@{}c@{}}Online \\ Adaptation\end{tabular} \\ \Xhline{2\arrayrulewidth}
Supervised & \redxmark                                                           & \redxmark                                                           & & \redxmark           & \redxmark                                                            \\
SSC-6D~\cite{peng2022self}     & \greencheckmark                                                           & \redxmark                                                           & & \redxmark           & \redxmark                                                            \\
RePoNet~\cite{fu2022category}  & \greencheckmark                                                           & \redxmark                                                           & & \redxmark           & \redxmark                                                            \\
UDA-COPE~\cite{lee2022uda}   & \greencheckmark                                                           & \redxmark                                                           & & \greencheckmark           & \redxmark                                                            \\
Self-DPDN~\cite{lin2022category}  & \greencheckmark                                                           & \greencheckmark                                                           & & \redxmark           & \redxmark                                                            \\ 
\cellcolor[rgb]{0.9,0.9,0.9}{Ours}       & \cellcolor[rgb]{0.9,0.9,0.9}{\greencheckmark}                                                           & \cellcolor[rgb]{0.9,0.9,0.9}{\greencheckmark}                                                           & \cellcolor[rgb]{0.9,0.9,0.9}{} & \cellcolor[rgb]{0.9,0.9,0.9}{\greencheckmark}            & \cellcolor[rgb]{0.9,0.9,0.9}{\greencheckmark}                                                           
\\ \Xhline{4\arrayrulewidth} 
\end{tabular}
}
    \label{tab:related_work}
\vspace{-0.5em}
\end{table}

In this paper, we propose Test-time Adaptation for Category-level Object Pose Estimation (\textbf{TTA-COPE}) to handle domain shifts without any target domain annotations (see \Fref{fig:teaser}).
Prior works on general test-time adaptation~\cite{wang2021tent, wang2022continual} propose self-training to minimize entropy loss.
TENT~\cite{wang2021tent} has shown improvement in 2D classification and segmentation tasks. 
We show, however, that simply extending TENT for the category-level object pose estimation is not effective.
Another self-training strategy is the teacher-student framework~\cite{tarvainen2017mean} with pseudo labels. 
However, since pseudo labels are created without any noise filtering, naive pseudo labels may be unreliable and cause convergence to a suboptimal model.

To tackle this problem, we design a novel pose ensemble method to perform  test-time adaptation for  category-level object pose estimation by extending the pose-aware filtering of UDA-COPE~\cite{lee2022uda}.
The proposed method uses an ensemble of teacher-student predictions based on pose-aware confidence, which is used both for generating pseudo labels and inference.
Also, the pose ensemble helps to train models with additional self-training loss to reduce the domain shift for category-level pose estimation by using pose-aware confidence. 
We demonstrate the advantages of our proposed pose ensemble and self-training loss with extensive studies in both semi-supervised and unsupervised settings.
We show that our TTA-COPE framework achieves state-of-the-art performance compared to strong TTA baselines.

In summary, the main contributions of our work are as follows:
\begin{itemize}
 \setlength\itemsep{-0.5em}
    \item We propose Test-Time Adaptation for Category-level Object Pose Estimation (TTA-COPE), which handles domain shifts without labeling target data and without accessing source data during test time.
    \item We introduce a pose ensemble with self-training loss that utilizes the teacher-student predictions to generate robust pseudo labels and estimates accurate poses for inference.
    \item We evaluate our framework with experimental comparisons against strong test-time baselines and state-of-the-art methods under both semi-supervised and unsupervised settings.
\end{itemize}

\section{Related Works}
\subsection{Supervised Methods}
Fully supervised learning methods for category-level object pose estimation~\cite{wang2019normalized, Tian2020prior, chen2020cass, chen2021sgpa, lin2021dualposenet, wang2021category} train their models using labeled source data (\eg, synthetic) and target data (\eg, real).
Most category-level 6D object pose and size estimation approaches~\cite{wang2019normalized, Tian2020prior, wang2021category, chen2021sgpa, lee2021category} use the dense correspondence via Normalized Object Coordinate Space (NOCS) representation as a common way to estimate pose and size.
These correspondence-based methods initially estimate the NOCS map from RGB or depth images. 
Afterwards, the Umeyama algorithm~\cite{umeyama1991least} with RANSAC is used to estimate optimal poses and object sizes by minimizing distances between depth and estimated NOCS map.

Some methods use category priors~\cite{Tian2020prior, chen2021sgpa, wang2021category} as the representative 3D shape per class, jointly reconstructing the full 3D shape and estimating the NOCS map from the full shape.
Results show that this prior category helps improve the accuracy of the NOCS map and enhance the pose estimation performance.
Other methods directly regress the pose or jointly utilize the correspondence representations~\cite{chen2020cass, lin2021dualposenet, irshad2022shapo}.

\subsection{Unsupervised Methods}
Given that annotating the 6D object pose and 3D size labels in the real world is expensive, time-consuming, and laborious, 
RePoNet~\cite{fu2022category} and SSC-6D~\cite{peng2022self} propose  semi-supervised approaches that reconstruct the entire shape and use differential rendering~\cite{liu2019soft, liu2020dist} techniques for self-training signals.
These are category-specific methods and utilize multiple models, as many models as the number of categories is required.
UDA-COPE~\cite{lee2022uda} proposes an unsupervised domain adaptation method to mitigate the domain shift from the source to the target domain with a single model to estimate all categories efficiently.

Although these works show reasonable performance without using pose labels, there is still a limitation in relying on 2D ground truth information (segmentation or bounding boxes) to train the segmentation network or pose network.
Self-DPDN~\cite{lin2022category} shows a fully unsupervised method using inter/intra-consistency as a reconstructed shape in a self-supervised objective but requires a supervised loss for the source domain in the unsupervised learning process.
Additionally, the self-training loss utilizes a full 3D shape and requires an additional  reconstruction module for 3D shape.
Most methods, except for UDA-COPE~\cite{lee2022uda}, jointly use the supervised loss using source label data during unsupervised learning to relax the unstable training. 
Furthermore, all the aforementioned unsupervised methods train their model in an offline manner and are unsuitable for online applications~\cite{mousavian20196, marder2016project}.

\subsection{Test-time Adaptation Methods}
Test-Time Adaptation (TTA) aims to enable the online adaptation of a pretrained model to the target domain without access to the source domain (source-free)~\cite{wang2021tent, niu2022efficient, song2023ecotta, wang2022continual}.
The source data is commonly inaccessible during inference time because it is inefficient to train on a huge amount of source data every iteration.
There may also be privacy or legal constraints to accessing source data~\cite{liu2021source}, thus TTA is a more difficult but practical task than UDA. 
From the point of view of real-world applications, it is necessary to adapt to the new scene in an online way.
Accordingly, test-time adaptation is necessary for the success of practical, real-world computer vision applications.
Wang~\etal propose Test entropy minimization (TENT)~\cite{wang2021tent}, which trains a network using a labeled source dataset, and adapts it to the unlabeled target dataset by updating the network parameters in batch norm layers using entropy loss. 
CoTTA~\cite{wang2022continual} proposes a continual test-time adaptation method on the 2D classification and semantic segmentation tasks, and it effectively reduces the error accumulations while continually changing target data.
Not limited to the 2D tasks, test-time adaptation methods have been applied to other tasks such as 3D segmentation~\cite{shin2022mm, prabhudesai2022test} and robot manipulation~\cite{mancini2018kitting}.

\begin{figure*}
\begin{center}
\vspace{-5pt}
\includegraphics[width=0.85\linewidth]{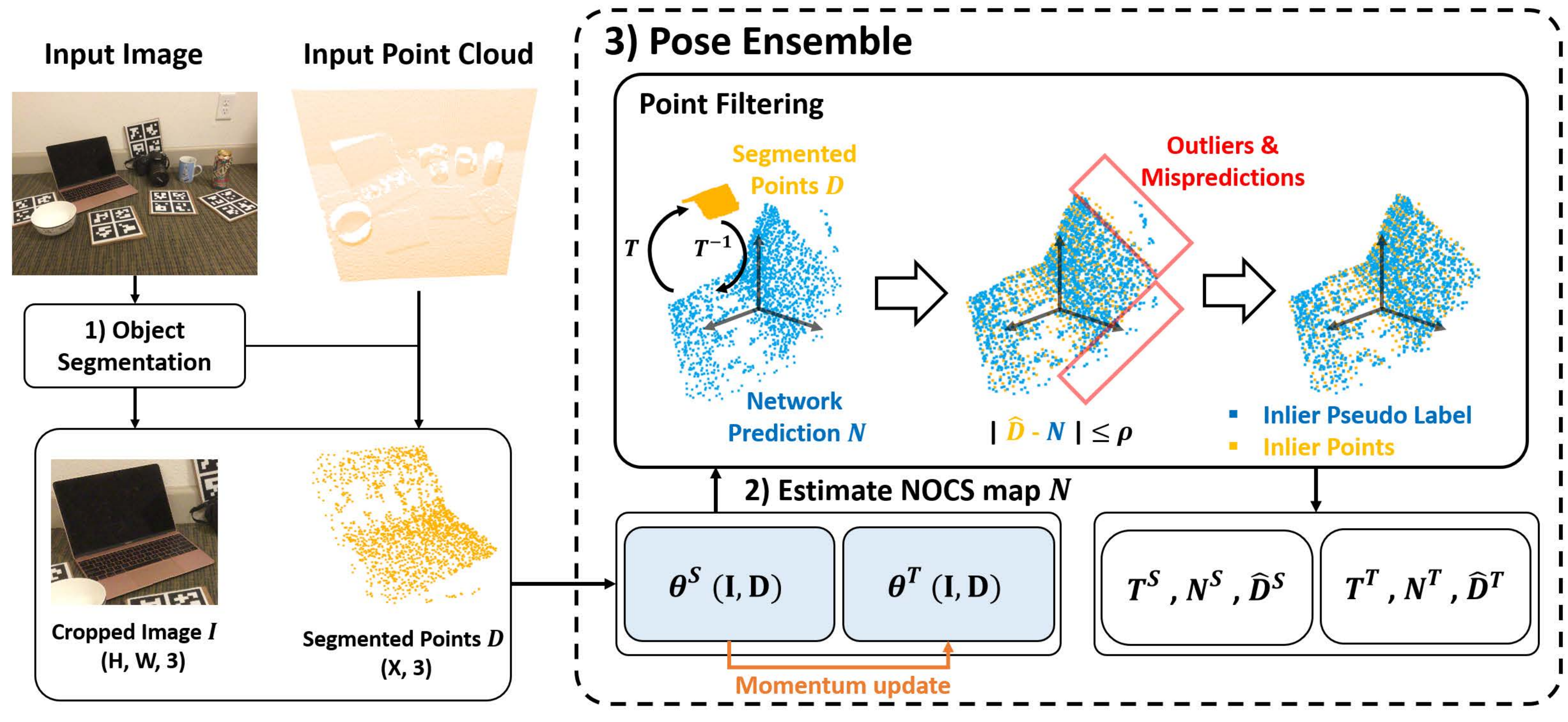}
\vspace{-0.075in}
\caption{\textbf{Overview of our Test-Time Adaptation for Category-level Object Pose Estimation (TTA-COPE) framework.}
Our method consists of three steps to estimate category-level object pose, similar to correspondence methods.
1) We detect and segment the object region using an off-the-shelf segmentation model from a single RGB image.
2) Given a cropped 2D object image $I$ and segmented 3D object point cloud $D$, our model estimates the correspondence points as a NOCS map $N$. 
3) Our proposed pose ensemble estimates the object pose $T$. 
It simultaneously utilizes the predictions of the student $\theta^S$ and teacher model $\theta^T$ by calculating the pose-aware confidence using the number of inlier points.
We operate point filtering for each student and teacher model and select a confident pose $T$ with higher confidence (\ie, more inlier points) for which the pose estimation is more accurate.
}
\vspace{-0.25in}
\label{fig:model}
\end{center}
\end{figure*}

\section{TTA-COPE}      
Given an RGB-D image, our approach aims to estimate the 6D pose $T\in \mathrm{SE}(3) $ and size $s \in \mathbb{R}^3_+$ of each object. 
The object pose $T$ is defined as the rigid transformation $[R\,|\,t]$, with rotation $R \in \mathrm{SO}(3)$ and translation $t \in \mathbb{R}^3$.

Our method consists of a two-stage learning scheme using source and target data, respectively.
In the first stage, we train the network using labeled (synthetic) source data in a supervised manner (\Sref{subsec:pretraining}). This step is the same as supervised methods and results in a pretrained model.
We then utilize this pretrained model for test-time adaptation (TTA) using unlabeled target data without accessing the source data.
To the best of our knowledge, we are the first to propose test-time adaptation for category-level object pose estimation.
Therefore, we study and explore how TTA baselines (\Sref{subsec:test_time_adaptation}) might apply to category-level object pose estimation. Finally, our proposed method is presented and explained (\Sref{subsec:pose_ensemble}).

\subsection{Pretraining with Source Data}\label{subsec:pretraining}
In this section, we introduce an overview of our model and then describe how to train the model using labeled source data.
We use UDA-COPE~\cite{lee2022uda} network as a base model, which we supplement with batch normalization (BN)~\cite{ioffe2015batch} in the 2D network to utilize BN updates for test-time adaptation.
\Fref{fig:model} shows an overview of our network, consisting of three steps to estimate the pose similar to correspondence methods~\cite{Tian2020prior, chen2021sgpa, lin2022category}.
First, we detect the bounding area of the object and segment the object surface using the off-the-shelf segmentation model from a single RGB image~\cite{he2017mask}.
In the second stage, given a cropped 2D object image $I$ and segmented 3D object point cloud $D$, our model estimates the correspondence as a NOCS map $N$.
We leverage the NOCS representation~\cite{wang2019normalized} to align diverse object instances within each class in a unified 3D space.
In the final stage, our proposed pose ensemble method estimates the object pose $T$. It simultaneously utilizes the predictions of the student and teacher model by calculating the pose confidence using the number of filtered points.

The network is pretrained in a fully supervised manner using labeled source data.
We minimize the predictions of the NOCS map $N$ using the cross-entropy loss $L_{CE}$ with ground truth labels $N^{GT}$.
We also use the consistency loss $L_{C}$ with 2D and 3D augmentation to make the network robust to noise in the input~\cite{lee2022uda}.
The supervised loss is formulated as: 
\begin{equation}
L_{sup} = \lambda_{CE} L_{CE}(N, N^{GT}) + \lambda_{C} L_{C}(N_{aug}, N),
\end{equation}
 where $N_{aug}$ is the estimated NOCS map from augmented 2D or 3D input, and $\lambda_{CE},\lambda_{C} \in \mathbb{R}_+$ are weights.

\subsection{Test-time Adaptation with Target Data}\label{subsec:test_time_adaptation}
The model is updated every iteration while simultaneously estimating the object pose of the current scene.
(Note that this is different from previous methods that update the model by running multiple epochs over the target data~\cite{lin2022category}.)

We have a design choice for the objective loss when updating the model during test time. 
One of the widely used test-time objective losses is the entropy loss proposed by TENT~\cite{wang2021tent},
\begin{equation}\label{eq:tent}
L_{ent} = -\sum p({x_t})\log{p({x_t})},
\end{equation}
where $p({x_t})$ is the probability of predictions $\theta^S({x_t})$ from target data $x_t$. 
This simple objective loss encourages sharp distributions by assigning the most probability.

Another common approach is utilizing pseudo labels from a teacher-student framework~\cite{tarvainen2017mean} with momentum update. 
The teacher model $\theta^{T}$ generates pseudo ground truth $\hat{y} = \theta^T(x_t)$, where $x_t=(I,D)$. 
The student model $\theta^S$ then uses the predictions of the teacher as the ground truth signals (pseudo labels) with cross-entropy loss,
\begin{equation}\label{eq:pl_loss}
L_{pl} = L_{CE} (\theta^{S}(x_t), \hat{y}).
\end{equation} 
After updating the student model $\theta^S_i$ $\xrightarrow{} \theta^S_{i+1}$ by minimizing \Eref{eq:pl_loss}, the teacher model $\theta^T_{i+1}$ is updated by momentum update, 
\begin{equation}\label{eq:momentum_update}
\theta^{T}_{i+1}  \leftarrow \gamma \theta^{T}_{i} + (1-\gamma)  \theta^{S}_{i+1},
\end{equation}
where $i$ stands for the time stamp of the iteration and $\gamma$ is the momentum smoothing factor.
The separation of the model structure mitigates some error accumulation by the pseudo label and momentum update.

Although these objective strategies from previous work potentially provide self-training signals for test-time adaptation, they become brittle when directly applied to a category-level object pose estimation task.
The pseudo labels $\hat{y}$ are still inherently exposed to noisy inputs and predictions, leading to errors in the student model $\theta^S$. 
Since $\theta^T$ cannot guarantee clean pseudo labels for the student, we need the ability to filter noise in the input or predictions and to yield more reliable pseudo labels to the $\theta^S$.

\subsection{Pose Ensemble}\label{subsec:pose_ensemble}
To solve this problem, we propose to extend the pseudo-label filtering proposed by UDA-COPE~\cite{lee2022uda} with pose ensemble processing.
We observe that the UDA-COPE framework and pose filtering, while showing notable results, is inefficient for two reasons.  First, the teacher model is only used for generating pseudo labels and not for inference.
Second, the student model is also only used for inference and has less influence in generating pseudo labels.

We propose instead to jointly utilize the teacher and student models for both generating pseudo labels and for inference, which we show in the experimental results produces a noticeable improvement.
To this end, we design a pose ensemble module to simultaneously use teacher and student model predictions.
\Fref{fig:model} shows an overview of the pose ensemble with point filtering.

The point filtering has three steps. Unlike UDA-COPE~\cite{lee2022uda}, we repeat these steps for both the student and teacher: 
1) We initially estimate $T^{S}$  using the Umeyama algorithm $\theta_{p}$ within the region of object points $D$ and estimated NOCS map $N^{S}$.
2) We transform observed depth to normalized object coordinate space $\widehat{D}^{S}$ by multiplying $(T^{S})^{-1}$ and $D$.
3) We compute each matching point distance between $N^{S}$ and $\widehat{D}$, removing outliers that exceed a certain threshold $\rho$.
This whole process of point-filtering is then repeated using $N^T$ to estimate $T^T$.  This process is represented as follows, where $N^{S,T}$ is either $N^S$ or $N^T$, and similarly for $T^{S,T}$:
\begin{equation} \label{eq:filtering}
\begin{split}
T^{S,T} &= \theta_{p}(D,N^{S,T}), \\
\widehat{D}^{S,T} &= \left(T^{S,T}\right)^{-1}D, \\
e^{S,T}_{j} &= \begin{cases}
            inlier      & \mbox{ if } \,  {\it \| \widehat{D}^{S,T}_j - N^{S,T}_j \|} \leq {\it \rho}, ~~ \forall j\\
            outlier   & \mbox{ otherwise}.    \\
            \end{cases}                             \\
\end{split}
\end{equation}
where $j=1,\ldots,X$, and $X=|D|=|{\widehat D}|$.

\begin{table*}
    \caption{\textbf{
    Quantitative comparisons with state-of-the-art methods on the REAL275 dataset.}
    }
  \vspace{-0.2in}
\centering
    \resizebox{0.97\linewidth}{!}{
\begin{tabular}{l cc cc cccccc}
\\ \Xhline{4\arrayrulewidth}  
\multirow{2}{*}{Method} & 
\multicolumn{2}{c}{Supervised} & 
\multicolumn{2}{c}{Unsupervised} & 
\multicolumn{6}{c}{mAP (↑)}                                       \\ 
\cmidrule(lr){2-3}
\cmidrule(lr){4-5}
                        & Source         & Target
                                 & Target & Online
                                                                                                                  & IoU$_{50}$ & IoU$_{75}$ & 5° 2cm & 5° 5cm & 10° 2cm & 10° 5 cm 
                        \\ \Xhline{2\arrayrulewidth}  
                        \multicolumn{11}{c}{\textbf{Source (Supervised)}}
                        \\

Metric Scale~\cite{lee2021category}               & 2D/3D                     &                                      &                                        &                                                                               & 54.6        & 8.4        & 2.2   & 5.4    & 10.1    & 25.0     \\
SPD~\cite{Tian2020prior}               & 2D/3D                     &                                      &                                        &                                                                               & 50.5        & 17.0        & 11.5   & 12.1    & 33.0    & 37.9     \\
\Xhline{2\arrayrulewidth}
\multicolumn{11}{c}{\textbf{Source (Supervised) \& Target (Supervised)}}                                                                          
\\ 
NOCS~\cite{wang2019normalized}                    & 2D/3D                    & 2D/3D                                    &                                        &                                                                               & 47.2        & 9.4        & 7.2    & 10.0    & 13.8    & 25.2     \\
SPD~\cite{Tian2020prior}             & 2D/3D                    & 2D/3D                                    &                                        &                                                                               & 68.5        & 27.0        & 19.5   & 21.6    & 43.5    & 54.0     \\
DualPoseNet~\cite{lin2021dualposenet}             & 2D/3D                    & 2D/3D                                    &                                        &                                                                               & ---         & 30.8        & 29.3   & 35.9    & 50.0    & 66.8     \\
CR-Net~\cite{wang2021category}                  & 2D/3D                    & 2D/3D                                    &                                        &                                                                               & ---        & 33.2         & 27.8   & 34.3    & 47.2    & 60.8     \\
SGPA~\cite{chen2021sgpa}                    & 2D/3D                    & 2D/3D                                    &                                        &                                                                               & ---          & 37.1        & 35.9   & 39.6    & 61.3    & 70.7     \\
\Xhline{2\arrayrulewidth}
\multicolumn{11}{c}{\textbf{Source (Supervised) \& Target (Semi-Supervised)}}
\\ 
SSC-6D~\cite{peng2022self}      & 2D/3D                    &       2D                               & 3D                                      &        \redxmark                                                                       & 73.0        &  ---       & 16.8   & 19.6    & 44.1    & 54.5     \\
RePoNet~\cite{fu2022category}     & 2D/3D                    &     2D                                 & 3D                                      &      \redxmark                                                                         & 76.0          &      ---       & 30.7   & 33.9    &      ---   & 63.0       \\
UDA-COPE~\cite{lee2022uda}      & 2D/3D                    &       2D                               & 3D                                      &        \redxmark                                                                       & 75.5        & 34.4        & 30.5   & 34.9    & 57.0    & 66.1     \\
Self-DPDN~\cite{lin2022category}     & 2D/3D                    &   2D                                   & 3D                                      &   \redxmark                                                                            & 75.2          & 41.6       & \textbf{39.5}   & \textbf{45.0}      & 63.3    & 72.2     \\

\cellcolor[rgb]{0.9,0.9,0.9}{TTA-COPE (Ours)}                    & \cellcolor[rgb]{0.9,0.9,0.9}{2D/3D}                    &      \cellcolor[rgb]{0.9,0.9,0.9}{2D}                                &  \cellcolor[rgb]{0.9,0.9,0.9}{3D}                                    &          \cellcolor[rgb]{0.9,0.9,0.9}{\greencheckmark}                                                                   & \cellcolor[rgb]{0.9,0.9,0.9}{\textbf{78.7}}        & \cellcolor[rgb]{0.9,0.9,0.9}{\textbf{43.5}}        & \cellcolor[rgb]{0.9,0.9,0.9}{33.3}   & \cellcolor[rgb]{0.9,0.9,0.9}{38.1}    & \cellcolor[rgb]{0.9,0.9,0.9}{\textbf{64.3}}    & \cellcolor[rgb]{0.9,0.9,0.9}{\textbf{75.1}}        \\
\Xhline{2\arrayrulewidth}
\multicolumn{11}{c}{\textbf{Source (Supervised) \& Target (Unsupervised)}}
\\
Self-DPDN~\cite{lin2022category}     & 2D/3D                    &                                      & 2D/3D                                      &          \redxmark                                                                     & 67.2          & \textbf{43.9}        & \textbf{39.0}   & \textbf{46.7}      & \textbf{61.8}    & \textbf{73.4}     \\
\cellcolor[rgb]{0.9,0.9,0.9}{TTA-COPE (Ours)}                    & \cellcolor[rgb]{0.9,0.9,0.9}{2D/3D}                    & \cellcolor[rgb]{0.9,0.9,0.9}{}                                      &     \cellcolor[rgb]{0.9,0.9,0.9}{2D/3D}                                 &                  \cellcolor[rgb]{0.9,0.9,0.9}{\greencheckmark}                                                            & \cellcolor[rgb]{0.9,0.9,0.9}{\textbf{69.1}}        & \cellcolor[rgb]{0.9,0.9,0.9}{39.7}        & \cellcolor[rgb]{0.9,0.9,0.9}{30.2}   & \cellcolor[rgb]{0.9,0.9,0.9}{35.9}    & \cellcolor[rgb]{0.9,0.9,0.9}{61.7}    & \cellcolor[rgb]{0.9,0.9,0.9}{73.2}        \\
\Xhline{4\arrayrulewidth}
\end{tabular}

\vspace{-0.2in}
}
 \label{tab:sota_table_all_full}
\end{table*}

Our proposed pose ensemble aims to fuse two predictions ($N^S, N^T$) to estimate the optimal pose.
Although this fusion could be conducted at the input level before obtaining the $T$ from $\theta_{p}$, we found that our proposal of fusing at the output level after getting $T$ produces slightly better results (see the ablation study, \Sref{subsec:ablation_study}).
We first estimate each pose ($T^S, T^T$) from each prediction ($N^S, N^T$) using $\theta_{p}$.
We then calculate the number of inliers ($inliers^S, inliers^T$) from each pose using \Eref{eq:filtering} to evaluate the confidence of the pose.
We assume that the higher this confidence (\ie, the more inliers) is, the more accurate the pose estimation will be. Therefore:
\begin{equation}\label{eq:pose_ensemble}
T_{out} = \begin{cases}
            T^S      & \mbox{ if } \,  {\it inliers}^S > {\it inliers}^T, \\
            T^T    & \mbox{ otherwise}.    \\
            \end{cases}                             \\
\end{equation}
We found that this simple pose-aware ensemble method is more effective for generating high-quality pseudo labels and inference than other ensemble methods (\Sref{subsec:ablation_study}).

Another crucial part of our method is a self-training loss. 
The total loss that we use to update the student network 
combines this self-training $L_d$ loss and the pseudo label loss $L_{pl}$:
\begin{equation}\label{eq:tta_loss}
\begin{split}
L_{tta} &= \lambda_{d}L_{d}  + \lambda_{pl}L_{pl}.
\end{split}
\end{equation}
The first term is self-training loss $L_d$, which helps learn the distribution of the target domain and reduce the domain gap through $T_{out}$, which is obtained from the pose ensemble and observed $D$.
This loss is given by:
\begin{equation}\label{eq:depth_loss}
L_{d} = L_{CE} (N^{S}_{e}, \text{Enc}(\widehat{D}_{e})), \\
\end{equation} 
where $N^S_e$ is the inlier-only student NOCS map, ${\widehat D}_e$ is the resulting inliner-only transformed depth point cloud in NOCS space, and $\text{Enc}(\cdot)$ is the one-hot encoding to enable cross-entropy loss.

The second term
\begin{equation}\label{eq:tta_losspl}
\begin{split}
L_{pl} &= L_{CE}(N^{S}_{e},N^{T}_{e})
\end{split}
\end{equation}
is the pseudo label loss from UDA-COPE~\cite{lee2022uda},
where $N^T_e$ is the inlier-only teacher NOCS map. Since $N^S=\theta^S(x_t)$, and ${\hat y}=\theta^T(x_t)$, $L_{pl}$ in \Eref{eq:tta_losspl} is the same as the loss in \Eref{eq:pl_loss}, except that here only inliers are considered.

\section{Experiments}
\subsection{Dataset}
We utilize two widely used category-level pose estimation datasets as the source and target domain, respectively. 
The {\em source data} is the Context-Aware MixEd ReAlity (CAMERA) dataset \cite{wang2019normalized}, generated by rendering and compositing synthetic objects into real scenes while considering the context.
CAMERA consists of 275K RGB-D images as a training set with 1,085 object instances chosen from six categories: bottle, bowl, camera, can, laptop and mug.
We use the REAL dataset \cite{wang2019normalized} as the {\em target domain}.
The target data consists of 4,300 real-world images of seven scenes for training and 2,750 real-world images of six scenes for evaluation.
We refer to REAL evaluation set as REAL275.
Our TTA methods do not use the target training set (4,300 images with seven scenes) and only use the evaluation set (2,750 images with six scenes, REAL275) for test-time adaptation, {\em e.g., TENT~\cite{wang2021tent}} process evaluation set in a sequential, online manner.
On the other hand, unsupervised methods such as~\cite{lin2022category, fu2022category, peng2022self} use the target training set for updating the models without any restrictions on how much and how long they use the information.

\subsection{Implementation Details}
\noindent{\textbf{Object Segmentation.}} We use Mask R-CNN~\cite{he2017mask} to obtain the object area in 2D image.
We use the identical results of Mask R-CNN for a fair comparison with previous methods~\cite{Tian2020prior, lin2021dualposenet, chen2021sgpa} for semi-supervised and unsupervised settings.
For the semi-supervised setting, Mask R-CNN has trained on the source, and target domain supervised manner.
For the unsupervised setting, Mask R-CNN is trained on only the source domain.
The detected area of the image resizes to 192 x 192 image resolution as the teacher or student model input.

\noindent{\textbf{Student and Teacher.}} The student and teacher models have identical design in 2D and 3D branches from UDA-COPE~\cite{lee2022uda}.
We utilize the PSPNet~\cite{zhao2017pyramid} with ResNet34~\cite{he2016deep} backbone for the 2D image feature extraction. 
For a 3D branch, we use the MinkowskiNet~\cite{choy20194d} and utilize sparse convolution operation with a 5cm voxel size.
Our NOCS representation uses the classification with 32 bins~\cite{wang2019normalized} instead of direct regression.
During the pretraining stage, we train our model on the source data for 50 epochs using the Adam optimizer by initializing the learning rate of 1e-4 with a batch size of 32.
The learning rate was reduced by a ratio of 0.6 (at 15k iterations), 0.3 (at 30k iterations), 0.1 (at 45k iterations), and 0.01 (at 60k iterations).
During test-time adaptation, our student model uses the same learning rate,  and the teacher smoothly updates using momentum update with $\gamma$ = 0.99.
Given target data, we first update the model every iteration and then estimate the pose.
We set $\lambda_\text{CE}$ = 1.0, $\lambda_\text{C}$ = 1e-6, $\lambda_\text{d}$ = 1.0, $\lambda_\text{pl}$ = 1.0 for our experiments.
The point filtering threshold $\rho$ was set to 0.05 for all experiments.

\noindent{\textbf{Metrics.}}
To evaluate the performance of 3D object detection and 6D pose estimation, we follow the previous pose and size evaluation metric from Wang~\etal~\cite{wang2019normalized}.
We report the mean average precision (mAP) at the 50$\%$ and  75$\%$ intersection over union (IoU) thresholds for 3D object detection.
We also report mAP for 6D object pose evaluation w.r.t. rotation and translation errors, where e.g.\ the 5° 5cm metric describes the percentage of pose predictions where the error is less than both 5° and 5cm and the same for other thresholds. We recalculated all the metrics with the improved code.

\begin{table}
    \caption{\textbf{Quantitative comparisons with TTA baselines for category-level object pose estimation on the REAL275 dataset.}}
\centering
    \resizebox{1.0\linewidth}{!}{
\begin{tabular}{l ccccc}
\Xhline{4\arrayrulewidth}
\multirow{2}{*}{Method} & \multicolumn{2}{c}{Unsupervised} & \multicolumn{3}{c}{mAP (↑)}                                       \\ \cline{2-3}
                                                      & Target 3D              & Target 2D             & IoU$_{75}$ & 5° 2cm & 5° 5 cm \\  \Xhline{2\arrayrulewidth}
TENT~\cite{wang2021tent} - \Eref{eq:tent}                                                & \checkmark       &                        & 33.7          & 25.9   & 29.1          \\
PL~\cite{tarvainen2017mean} - \Erefs{eq:pl_loss}{eq:momentum_update}                                                  & \checkmark       &                      & 39.9        & 29.7   & 34.9     \\
PL-F~\cite{lee2022uda} - \Erefs{eq:pl_loss}{eq:filtering}                                                 & \checkmark       &                         & 41.1        & 31.4   & 36.3    \\
\cellcolor[rgb]{0.9,0.9,0.9}{TTA-COPE (Ours)}                                              & \cellcolor[rgb]{0.9,0.9,0.9}{\checkmark}       &     \cellcolor[rgb]{0.9,0.9,0.9}{}                    & \cellcolor[rgb]{0.9,0.9,0.9}{\textbf{43.5}}        & \cellcolor[rgb]{0.9,0.9,0.9}{\textbf{33.3}}   & \cellcolor[rgb]{0.9,0.9,0.9}{\textbf{38.1}}    \\ \hline
TENT~\cite{wang2021tent} - \Eref{eq:tent}                                                & \checkmark       & \checkmark              & 32.3        & 26.8   & 31.2     \\
PL~\cite{tarvainen2017mean} - \Erefs{eq:pl_loss}{eq:momentum_update}                                                  & \checkmark       & \checkmark             & 36.0        & 26.9   & 33.1    \\
PL-F~\cite{lee2022uda} - \Erefs{eq:pl_loss}{eq:filtering}                                               & \checkmark       & \checkmark             & 36.5        & 28.0   & 34.0     \\
\cellcolor[rgb]{0.9,0.9,0.9}{TTA-COPE (Ours)}                                                & \cellcolor[rgb]{0.9,0.9,0.9}{\checkmark}       & \cellcolor[rgb]{0.9,0.9,0.9}{\checkmark}        & \cellcolor[rgb]{0.9,0.9,0.9}{\textbf{39.7}}        & \cellcolor[rgb]{0.9,0.9,0.9}{\textbf{30.2}}   & \cellcolor[rgb]{0.9,0.9,0.9}{\textbf{35.9}}      
\\
\Xhline{4\arrayrulewidth}
\end{tabular}
}
\label{tab:tta_comparison}
 \vspace{-0.15in}
\end{table}

\begin{table*}
    \caption{\textbf{Ablation study on variants of pose ensemble methods and self-training loss under the semi-supervised setting.}}
\centering
 \vspace{-6pt}
    \resizebox{0.99\linewidth}{!}{
    \setlength{\tabcolsep}{4pt}
\begin{tabular}{c cccc c cccccc}
\Xhline{4\arrayrulewidth}
\multirow{2}{*}{Method} & \multicolumn{5}{c}{Pose Ensemble}           & \multicolumn{6}{c}{mAP (↑)} \\ \cmidrule(lr){2-6}
& Input & Output  & Inference   & Pseudo Label  & $L_d$ (\ref{eq:depth_loss})                                                                     & IoU$_{50}$ & IoU$_{75}$ & 5° 2cm & 5° 5 cm & 10° 2cm & 10° 5 cm 
\\ \Xhline{2\arrayrulewidth}
LB &  &         &      &           &                                                                      & 76.2        & 37.5        & 29.1     & 34.6    & 62.1    & 73.2     \\ \hline
(1) & Argmax Match &         & \checkmark     &  &                                                                      & 71.1        & 27.5        & 29.6   & 34.3    & 60.7    & 71.7     \\
(2) & Softmax Avg.  &         & \checkmark     &  &                                                                      & \textbf{78.7}        & 40.4        & 32.1   & 37.3    & 64.0    & 74.7     \\
(3) & Softmax Max   &         & \checkmark     &  &                                                                      & 77.6        & 40.9        & 31.6   & 36.4    & 63.4    & 74.3     \\
(4) &      & Softmax Max     & \checkmark     &  &                                                                      & 77.5         & 41.1        & 32.1   & 37.1    & 64.3    & 74.6     \\
(5) &      & Inliers Max & \checkmark     &  &                                                                      & 77.6         & 41.2        & 32.8   & 37.6    & 64.4    & 74.9     \\
(6) &      & Inliers Max & \checkmark     &   & \checkmark                                                             & 78.2         & 42.9        & 32.4   & 37.8    & 64.1    & 74.7     \\
(7) & Argmax Match &         &              & \checkmark           &                                                                      & 78.3        & 40.7        & 31.7   & 36.4      & 63.2    & 73.9     \\
(8) & Softmax Avg.  &         &              & \checkmark           &                                                                      & 78.6        & 41.1        & 31.5     & 36.2    & 63.6    & 74.5     \\
(9) & Softmax Max   &         &              & \checkmark           &                                                                      & 78.1        & 41.2        & 31.9   & 36.5    & 63.5    & 74.4     \\
(10) &      & Softmax Max     &              & \checkmark           &                                                                      & 78.6        & 41.2        & 31.8   & 36.3    & 63.5    & 74.4     \\
(11) &      & Inliers Max &              & \checkmark           &                                                                      & 78.2        & 41.3        & 31.6   & 36.2    & 63.6    & 74.3     \\
(12) &      & Inliers Max &              & \checkmark           & \checkmark                                                             & \textbf{78.7}        & 43.0          & 32.6     & 37.0    & 63.5    & 73.9     \\
\cellcolor[rgb]{0.9,0.9,0.9}{Ours} &   \cellcolor[rgb]{0.9,0.9,0.9}{}   & \cellcolor[rgb]{0.9,0.9,0.9}{Inliers Max} & \cellcolor[rgb]{0.9,0.9,0.9}{\checkmark}     & \cellcolor[rgb]{0.9,0.9,0.9}{\checkmark}  & \cellcolor[rgb]{0.9,0.9,0.9}{\checkmark}                                                             & \cellcolor[rgb]{0.9,0.9,0.9}{\textbf{78.7}}        & \cellcolor[rgb]{0.9,0.9,0.9}{\textbf{43.5}}        & \cellcolor[rgb]{0.9,0.9,0.9}{\textbf{33.3}}   & \cellcolor[rgb]{0.9,0.9,0.9}{\textbf{38.1}}    & \cellcolor[rgb]{0.9,0.9,0.9}{\textbf{64.6}}    & \cellcolor[rgb]{0.9,0.9,0.9}{\textbf{75.1}}    
 \\ \Xhline{4\arrayrulewidth}
\end{tabular}
}
\label{tab:ablation_study}
\vspace{-0.15in}
\end{table*}

\subsection{Comparison with state-of-the-art}\label{subsec:sota_comparison}
\Tref{tab:sota_table_all_full} summarizes quantitative results on the REAL275 dataset under different settings:
1) supervised only on source data, 2) supervised on source and target data, 3) supervised on source data and semi-supervised on target data, and 4) supervised on source data and unsupervised on target data.
Not surprisingly, methods following 2) such as SPD~\cite{Tian2020prior} perform better than setting 1) since they have the benefit of accessing both source and target data.
Recent unsupervised domain adaptation (UDA) methods~\cite{lin2022category, lee2022uda} show remarkable results compared to state-of-the-art supervised methods without using target 3D labels (semi-supervised setting).
Our method shows state-of-the-art results in IoU metrics under the semi-supervised setting and even outperforms the recent supervised methods SGPA~\cite{chen2021sgpa} by a large margin (6.4 mAP in IoU$_{75}$).
Our TTA-COPE uses less time and data to train the target domain because of the advantage of test-time adatpation but achieves comparable results to the SOTA method, Self-DPDN~\cite{lin2022category}. 
Ours takes about 31 minutes (\Tref{tab:ablation_interval}) for TTA and is 58x faster than the Self-DPDN, which takes about 30 hours for training in the target domain, excluding inference time.

\subsection{Comparison with TTA baselines}\label{subsec:tta_baselines}
\Tref{tab:tta_comparison} summarizes results for the different design test-time adaptation (TTA) baselines in semi-supervised and unsupervised settings.
TENT~\cite{wang2021tent} has been designed for general unsupervised settings, but it is not specially designed for the object pose estimation task and naturally observes overall poor performance compared to other TTA baselines.
Mean teacher with pseudo labels (PL)~\cite{tarvainen2017mean} performs better than TENT~\cite{wang2021tent} in semi-supervised and unsupervised settings, and we believe that training the model using pseudo labels with momentum update (\Erefs{eq:pl_loss}{eq:momentum_update}) provides more stable training signals by reducing the error accumulation than entropy minimization (\ref{eq:tent}).
However, using pseudo labels without filtering (PL) provides unreliable labels as the ground truth, we found that pseudo label filtering (PL-F)~\cite{lee2022uda} performs better than PL, which indicates that noise filtering (\Eref{eq:filtering}) improves the results for the student as expected.
Finally, our TTA-COPE achieves state-of-the-art (SOTA) performance among all TTA baselines under both semi-supervised and unsupervised settings.

\begin{figure*}
\vspace{-8.0mm}
\begin{center}
\includegraphics[width=0.98\linewidth]{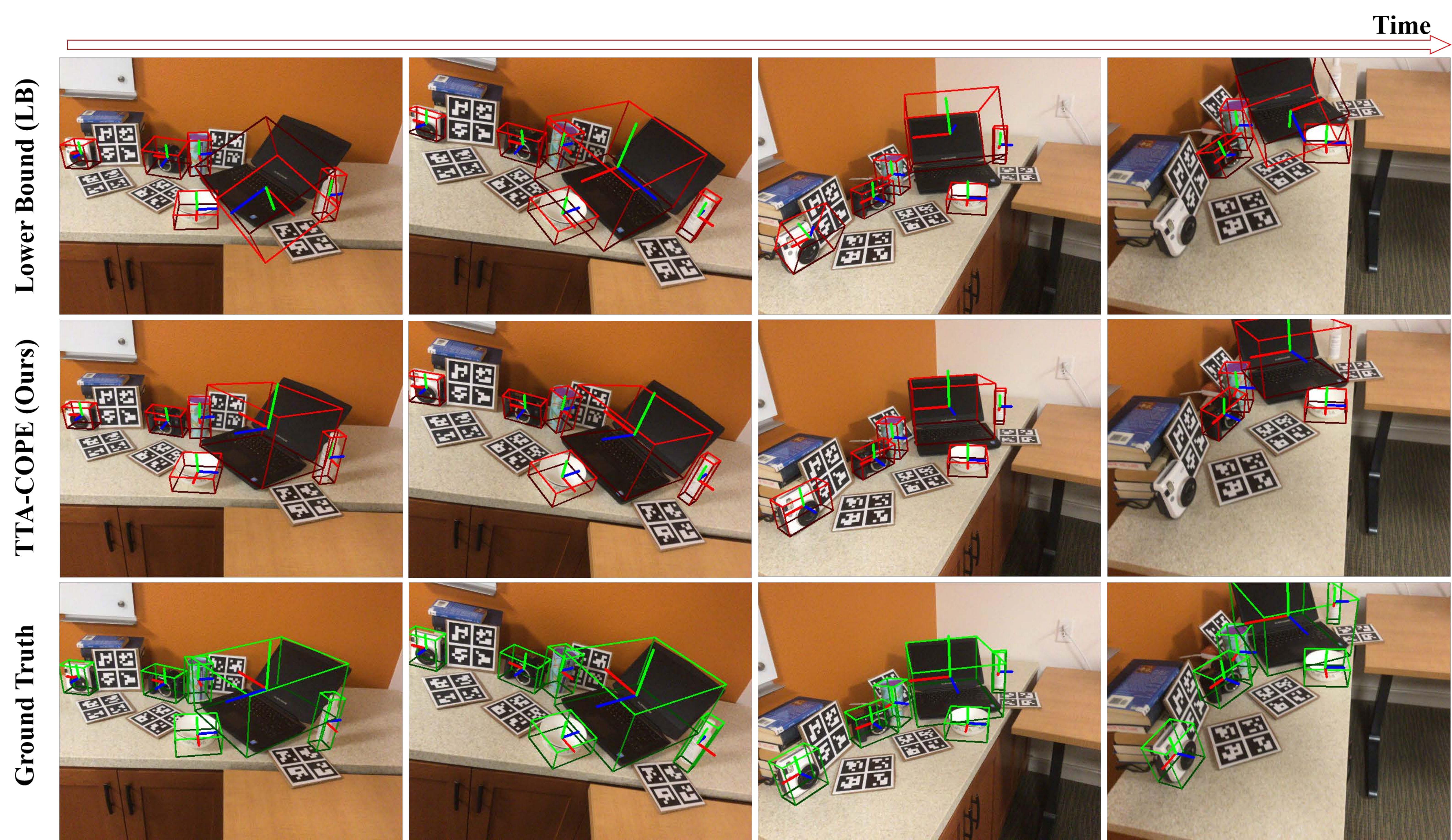}
\caption{
\textbf{Qualitative comparison with the lower bound (LB) and our TTA-COPE under the semi-supervised setting.}}
\vspace{-0.1in}
\label{fig:qualtiative_lower_bound_tta}
\end{center}
\vspace{-0.13in}
\end{figure*}

\subsection{Ablation Study}\label{subsec:ablation_study}
In this section, we conduct experiments to evaluate the efficacy of the pose ensemble and self-training loss under a semi-supervised setting.

\noindent{\textbf{Input/Output-Level Pose Ensemble.}} As mentioned in the \Sref{subsec:pose_ensemble}, we compare the different pose ensemble methods given predictions of teacher and student models ($N^T$, $N^S$) and answer the following question: Which ensemble is the most effective for test-time adaptation?
We categorize two ensemble techniques: 

\noindent 1) Input-level ensemble that fuses two NOCS map predictions ($N^T$, $N^S$) and makes an accurate single NOCS map $N$ to estimate poses using the Umeyama algorithm $\theta_p$. For the input-level ensemble that fuses two predictions ($N^T$, $N^S$) into one prediction $N$, we compare three strategies: Argmax Match, Softmax Average, and Softmax Max operations~\cite{jaritz2020xmuda, shin2022mm}.

\noindent 2) We ensemble output-level predictions ($N^T$, $N^S$) to estimate each pose ($T^T, T^S$) and choose the best pose predictions under specific criteria. Unlike the input ensemble, only the Softmax Max operation is valid for the output-level ensemble due to the nonlinearity of poses.

Table~\ref{tab:ablation_study} summarizes our ensemble ablation study results.
Table~\ref{tab:ablation_study}-(1-3) shows that the Softmax Avg. and Softmax Max perform better than the Argmax Match operation among input-level pose ensembles.
If we compare the same operation (Softmax Max) in the input-level, Table~\ref{tab:ablation_study}-(3), and output-level, Table~\ref{tab:ablation_study}-(4), the output-level ensemble shows slightly better performance since the output-level operation is directly related to the pose results.
Our proposed pose ensemble, Table~\ref{tab:ablation_study}-(5), considers pose-aware confidence by choosing a higher number of inlier points and achieves the best performance in all pose ensembles in Table~\ref{tab:ablation_study}-(1-5).

\noindent{\textbf{Use of Pose Ensemble.}} 
Our pose ensemble can be used for inference Table~\ref{tab:ablation_study}-(1-6, Ours) as well as generating pseudo labels Table~\ref{tab:ablation_study}-(7-12, Ours).
The results show that our pose ensemble is helpful for both cases. 
In particular, applying our ensemble to inferences Table~\ref{tab:ablation_study}-(5) has been shown to be more effective than application to generating pseudo labels Table~\ref{tab:ablation_study}-(11) in pose metric. 
We also observe that our pose ensemble used for inference or pseudo labels improves overall metrics compared to the lower bound (LB) that is only trained on source data without test-time adaptation.

\noindent{\textbf{Effect of self-training loss.}} The results of Table~\ref{tab:ablation_study}-(6) and Table~\ref{tab:ablation_study}-(12) show the effect of self-training loss~\Eref{eq:depth_loss}.
It yields an improvement of more than 1.7 mAP compared to without self-training loss results of Table~\ref{tab:ablation_study}-(5, 11) in the IoU$_{75}$ metric.
This indicates that the self-learning loss helps to learn the distribution of the target domain and reduces the domain shift by comparing the predicted NOCS map against the observed point cloud in the target domain.
The result in the last column is our proposed model for test-time adaptation, \Fref{fig:qualtiative_lower_bound_tta} shows qualitative results against the lower bound method.

\begin{table}
\caption{\textbf{Ablation study on different updating intervals.}}
\vspace{-3.0mm}
\centering
\resizebox{1.0\linewidth}{!}{
\begin{tabular}{l cccc c} 
\Xhline{4\arrayrulewidth}
Method & TTA Time & IoU$_{75}$ & 5° 2cm & 5° 5 cm \\ \Xhline{2\arrayrulewidth}
\cellcolor[rgb]{0.9,0.9,0.9}{Interval=1 (Ours)} & \cellcolor[rgb]{0.9,0.9,0.9}{31 min}     & \cellcolor[rgb]{0.9,0.9,0.9}{43.5}        & \cellcolor[rgb]{0.9,0.9,0.9}{33.3}   & \cellcolor[rgb]{0.9,0.9,0.9}{38.1}   \\ 
Interval=10 & 16 min & 41.8        & 32.7   & 37.7   \\  
Interval=20 & 15 min & 41.1        & 31.9   & 37.1   \\  
\Xhline{4\arrayrulewidth}
\end{tabular}
}
\label{tab:ablation_interval}
\vspace{-5mm}
\end{table}

\noindent{\textbf{Updating Interval.}}
TTA is inevitably slower than simple inference since additional training is required.
However, reducing the update interval enables faster TTA time than TTA baselines because of reduced training time.
Table~\ref{tab:ablation_interval} shows the difference in TTA speed and performance according to the updating interval.
Specifically, the interval of every 10 frames improves the speed of TTA roughly two-fold compared to Interval 1 (Ours), with a marginal performance drop.
We also increase intervals to 20, but it does not show as much improvement as before since most of the bottleneck arises from inference time.

\section{Conclusion}
We have proposed TTA-COPE, a test-time adaptation method for category-level object pose estimation that addresses the source-to-target domain shift without accessing source data at test time and without labeled target data.
Specifically, we designed a pose ensemble method with self-training for test-time adaptation that simultaneously uses the teacher-student model to generate robust pseudo labels and estimate accurate poses for inference.
We explore limitations of several test-time adaptation baselines and show that the proposed method achieves state-of-the-art performance. 
We demonstrate the benefits of our proposed pose ensemble and self-training loss with extensive studies in both semi-supervised and unsupervised settings.

To the best of our knowledge, TTA-COPE is the first approach that tries to solve test-time adaptation for category-level object pose estimation.
Since our method currently focuses on generating 6D pose labels, it does not affect 2D labels and the segmentation model. 
In future work, when jointly considering 2D label and Mask R-CNN, greater performance improvement and stable test-time adaptation would be possible.
Also, our pose estimation relies on non-differentiable pose estimation, and as such we could benefit from a differentiable pose estimation method~\cite{pineda2022theseus}.

\section*{Acknowledgment}
This work was part of Taeyeop Lee’s internship at NVIDIA and was also partially supported by an Institute of Information \& communications Technology Planning \& Evaluation(IITP) grant funded by the Korean government(MSIT) (No.2020-0-00440).

{\small
\bibliographystyle{ieee_fullname}
\bibliography{egbib}
}

\end{document}